\documentclass[10pt,twocolumn,letterpaper]{article}
\usepackage{authblk}

\usepackage[pagenumbers]{cvpr} % To force page numbers, e.g. for an arXiv version
\usepackage[utf8]{inputenc} % allow utf-8 input
\usepackage[T1]{fontenc}    % use 8-bit T1 fonts
\usepackage{url}            % simple URL typesetting
\usepackage{booktabs}       % professional-quality tables
\usepackage{amsfonts}       % blackboard math symbols
\usepackage{nicefrac}       % compact symbols for 1/2, etc.
\usepackage{microtype}      % microtypography
\usepackage{xcolor}         % colors

\usepackage{graphicx}       
\usepackage{amsmath}
\usepackage{multirow}
\usepackage{tabularx}
\usepackage{tabularray}
\usepackage{subcaption}
\usepackage{caption}
\usepackage{array}
\newcolumntype{C}{>{\centering \arraybackslash}X}
\usepackage{wrapfig}
\usepackage{xspace}

\newlength\savewidth
\newcommand{\tablestyle}[2]{\setlength{\tabcolsep}{#1}\renewcommand{\arraystretch}{#2}\centering\footnotesize}

\definecolor{cvprblue}{rgb}{0.21,0.49,0.74}
\usepackage[pagebackref,breaklinks,colorlinks,citecolor=cvprblue]{hyperref}

\title{SOAR: Self-Occluded Avatar Recovery \\from a Single Video In the Wild}

\author[1,2,*]{Zhuoyang Pan}
\author[1]{Angjoo Kanazawa}
\author[1,*]{Hang Gao}
\affil[1]{UC Berkeley}
\affil[2]{ShanghaiTech University}

\begin{document}

\twocolumn[
{%
\renewcommand\twocolumn[1][]{#1}%
\maketitle
\vspace{-3em}
\begin{center}
\includegraphics[width=\textwidth]{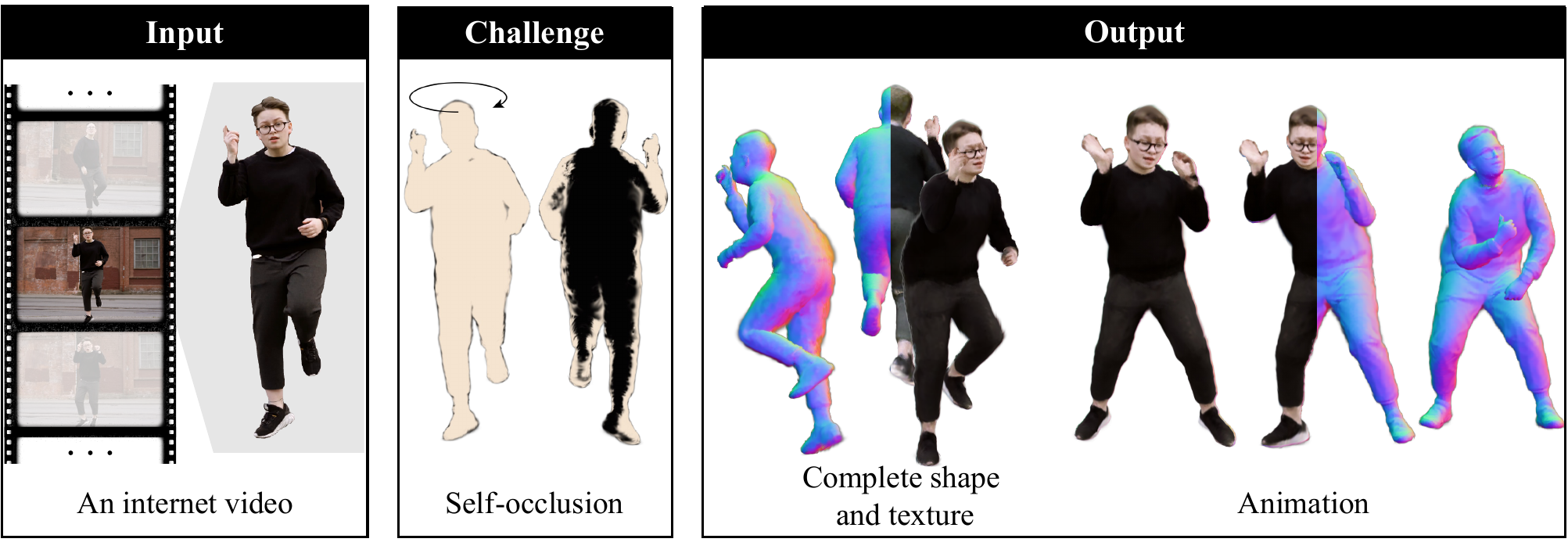}
\vspace{-2em}
\captionof{figure}{
    \textbf{Complete human reconstruction from partial observations in the wild.}
    We present \textbf{SOAR}: \textbf{S}elf-\textbf{O}ccluded \textbf{A}vatar \textbf{R}ecovery. 
    Given a video of a moving human where parts of the body are entirely unobserved (left), SOAR recovers a photo-realistic avatar with complete texture and shape (right), by leveraging structural human normal prior and generative diffusion prior.
}
\label{fig:teaser}
\end{center}%
}]
\renewcommand{\thefootnote}{}
\footnotetext{* Equal Contribution.}

\begin{abstract}
\vspace{-0.8em}
Self-occlusion is common when capturing people in the wild, where the performer do not follow predefined motion scripts. 
This challenges existing monocular human reconstruction systems that assume full body visibility. 
We introduce Self-Occluded Avatar Recovery (SOAR), a method for complete human reconstruction from partial observations where parts of the body are entirely unobserved. 
SOAR leverages structural normal prior and generative diffusion prior to address such an ill-posed reconstruction problem. 
For structural normal prior, we model human with an reposable surfel model with well-defined and easily readable shapes. 
For generative diffusion prior, we perform an initial reconstruction and refine it using score distillation.
On various benchmarks, we show that SOAR performs favorably than state-of-the-art reconstruction and generation methods, and on-par comparing to concurrent works. Additional video results and code are available at \url{https://soar-avatar.github.io/}.

\end{abstract}

\vspace{-1em}
\section{Introduction}
% \looseness=-1
Recovering life-like human avatar from a single in-the-wild video, such as internet footage or smartphone capture, is crucial for advancing virtual reality, robotics, and content creation. 
This task is challenging due to dynamic modeling and the lack of effective multi-view signals~\cite{gao2022monocular}.
Despite tremendous progress~\cite{peng2021neural,weng2022humannerf,li2022tava,jiang2023instantavatar,zielonka2023drivable,zheng2024physavatar} in recent years, success in human reconstruction methods in the wild remains limited. 
One key reason is that existing approaches often assume full visibility of the human body, which fails in most of unscripted casual captures.
For this ill-posed problem, reconstruction alone is insufficient.

We present SOAR, a general system for human avatar recovery from a single self-occluded video in the wild.
In Figure~\ref{fig:teaser}, we demonstrate our setting and results.
We tackle this challenging problem with two key insights.
First, to optimize with ill constraints, we need stronger data terms and more parsimonious representations.
Second, we need to combine reconstruction with generation based on how many observations we have. 
With more observations, we prioritize reconstruction to preserve details like identity. 
With fewer observations, generation becomes crucial. 
A successful system should seamlessly integrate these two components.

Motivated by these two insights, we model the human avatar as a globally consistent set of Gaussian surfels~\cite{dai2024high} with well-defined and easily readable normals. 
We model articulation between different poses using a simple forward mapping with linear blend skinning~\cite{SMPL:2015}.
We fit this compact, dynamic human representation to a general self-occluded video in the wild by incorporating two additional sources of supervision on top of the input RGB data: structural human normal prior~\cite{xiu2022icon, xiu2023econ} and generative diffusion prior~\cite{wang2023imagedream}. 
They provide strong shape and texture constraints for unobserved regions, crucial in our challenging problem setup.
% Our model estimates visibility on inferred geometry by fusing information across the entire video sequence.
To this end, our approach is able to recover complete photo-realistic avatar with highly detailed geometry, which can be used for real-time rendering and animation.

To investigate the effectiveness and robustness of our approach, we compare against reconstruction-based~\cite{lei2023gart,hu2024gaussianavatar} and generation-based approaches~\cite{ho2023sith} as baselines.
We also compare with concurrent work HAVE-FUN~\cite{yang2024have} that reconstructs from partial observations on its own experimental protocols using the official open-source implementation.
Extensive experiments show that SOAR performs favorably than state-of-the-art reconstruction and generation methods, and on-par comparing to concurrent works.

\section{Related work}
\subsection{3D Gaussian and surfel splatting}
\begin{figure}
    \centering
    \includegraphics[width=\linewidth]{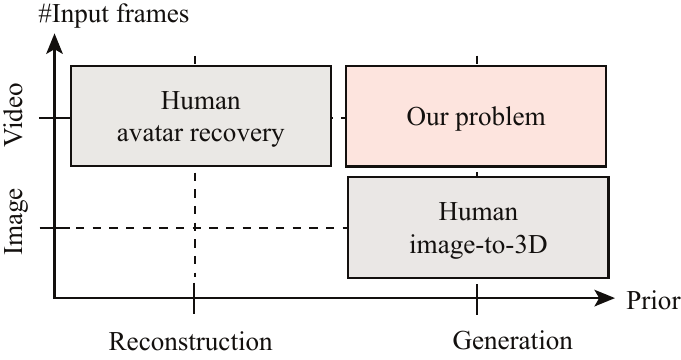}
    \caption{\small\textbf{Relation to existing problems.}
    Our problem requires combining human reconstruction from video frames and human generation for occluded regions.
    }
    \label{fig:related}
\end{figure}
Neural rendering has advanced significantly since the introduction of NeRF~\cite{mildenhall2021nerf}. 3D Gaussian Splatting~\cite{kerbl3Dgaussians} is particularly notable for its efficiency in high-resolution synthesis and real-time rendering.
It represents scenes as explicit 3D Gaussians, allowing direct rasterization in pixel space that is much faster than volume integration.
However, 3D Gaussians struggle with accurate scene geometry recovery.
Various attempts~\cite{guedon2023sugar,jiang2023gaussianshader,liang2023gs} have been made to solve this problem.
Recently, 2D Gaussian Splatting~\cite{huang20242d} and Gaussian surfels~\cite{dai2024high} propose to flatten 3D Gaussians into surfels, making geometry easier to readout and coupled with RGB rendering.
Our work builds on these advancements and use surfel model for precise human shape recovery while preserving effective appearance modeling.
\subsection{Neural rendering for human reconstruction}
\looseness=-1
Neural rendering significantly advances template-based human reconstruction~\cite{carranza2003free,de2008performance,gall2009motion,stoll2010video} by allowing 3D avatar recovery from 2D images.
Recent works have focused on dynamic modeling, out-of-distribution reposing, and runtime efficiency.
NeuralBody~\cite{peng2021neural} and Vid2Avatar~\cite{guo2023vid2avatar} are canonical frameworks in the first category. 
For reposing, Animatable NeRF~\cite{peng2021animatable}, TAVA~\cite{li2022tava}, and InstantAvatar~\cite{jiang2023instantavatar} use inverse blend skinning or root finding~\cite{chen2021snarf} to ensure consistency. 
Recent methods~\cite{zielonka2023drivable,svitov2024haha} employ 3D Gaussians for efficient rendering. 
We select GART~\cite{lei2023gart} and GaussianAvatar~\cite{hu2024gaussianavatar} as baseline in our experiments.
We also compare with concurrent work HAVE-FUN~\cite{yang2024have} that aims to recover complete avatar from partial observations on its own benchmarks.
We found that existing benchmarks all assume full-body visibility, even for our concurrent works, and thus test our methods on a new evaluation split from DNA-Rendering~\cite{cheng2023dna}.

\subsection{Diffusion prior for human generation}
Score distillation sampling~\cite{poole2022dreamfusion} has shown that 2D diffusion models are effective 3D priors for content creation.
Since then, significant progress has been made in making 3D generation more stable~\cite{shi2023mvdream,wang2023imagedream}, realistic~\cite{wang2024prolificdreamer} and efficient~\cite{yi2023gaussiandreamer,tang2023dreamgaussian}.
In the human modeling community, this paradigm has also been adopted, with notable works~\cite{kolotouros2024dreamhuman,liao2023tada,yuan2023gavatar} incorporating predefined SMPL templates~\cite{SMPL:2015} to bias the generation process.
Works on human image-to-3D~\cite{huang2023tech,zhang2023sifu,ho2023sith} aim to recover human avatars from a single photo. 
For example, TeCH~\cite{huang2023tech} optimizes a differentiable tetrahedron representation~\cite{shen2021dmtet} through score distillation, while SiTH~\cite{ho2023sith} directly optimizes an SDF field from diffused images. 
We include SiTH as a baseline in our work.
However, these approaches struggle with video input, resulting in temporally inconsistent prediction when applied frame-by-frame, investigated in Section~\ref{sec:dna-rendering}.
Our work fuses pose-conditioned, noisy diffusion priors into a single, globally consistent avatar model.

\section{Method}
We aim to recover photo-realistic human avatar from a single self-occluded in-the-wild video, where parts of the human body remain unobserved. 
This highly ill-posed problem necessitates stronger priors and better avatar representation.
\begin{figure*}
    \centering
    \includegraphics[width=\linewidth]{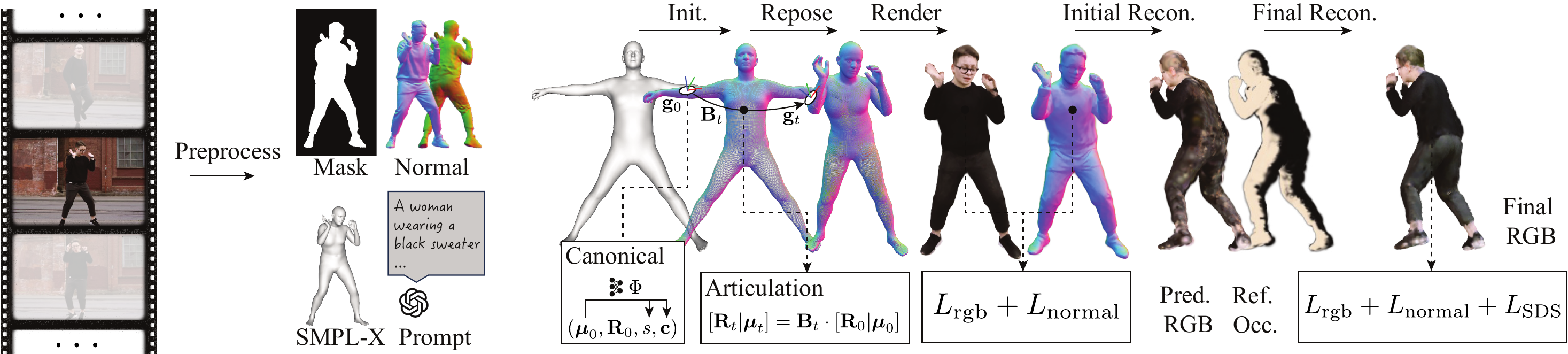}
    \caption{
    \textbf{System overview.}
    Given an input video, we preprocess for frame-wise mask, front and back normal, SMPL-X parameters, as well as video-level text prompt description (Section~\ref{sec:preprocessing}).
    Our model consists of a canonical Gaussian surfel representation and an articulation representation (Section~\ref{sec:avatar}).
    We perform initial reconstruction while estimating occlusion, producing partially completed avatar due to the lack of observation (Section~\ref{sec:reconstruction}), which is then refined by generative diffusion priors (Section~\ref{sec:generation}).
    }
    \label{fig:system}
\end{figure*}

Existing human reconstruction methods~\cite{peng2021neural,guo2023vid2avatar,li2023animatable,li2022tava,jiang2023instantavatar,lei2023gart,hu2024gaussianavatar} require the performer to reveal 360 views of their body, which does not often occur in internet videos.
Conversely, existing human image-to-3D methods~\cite{huang2023tech,zhang2023sifu,ho2023sith,albahar2023single} can only condition on one input view, producing inconsistent results across frames.
Our method bridges the gap between reconstruction and generation, addressing these challenges to produce consistent and accurate human avatars with self-occlusion.

The rest of this section is organized as follows.
First, we talk about the preprocessing step given a single in-the-wild video (Section~\ref{sec:preprocessing})
Next, we discuss our avatar model, represented as a globally consistent set of 3D Gaussian surfels~\cite{dai2024high} that transforms from a canonical space to each pose configuration (Section~\ref{sec:avatar}).
Then, we fuse RGB and structural normal supervision through an initial reconstruction while estimating occlusion in 3D~(Section~\ref{sec:reconstruction}).
Finally, we refine this initial reconstruction using score distillation (Section~\ref{sec:generation}).
Our whole pipeline is illustrated in Figure~\ref{fig:system}.

% \vspace{-0.1em}
\subsection{Preprocessing}
\label{sec:preprocessing}
Given a sequence of video frames capturing a moving person, we prepare a set of estimates using off-the-shelf methods.
Specifically, for each frame $\mathbf{I}_t$, we estimate the foreground mask $\mathbf{M}_t$ using SAM~\cite{kirillov2023segment}, generate a video-level text prompt $\mathbf{p}$ using GPT-4o~\cite{achiam2023gpt}, obtain front and back normal maps $(\mathbf{N}_t, {}^B\mathbf{N}_t)$ using ICON~\cite{xiu2022icon}, and infer 2D keypoints $\mathbf{k}_t \in \mathbb{R}^{137 \times 2}$ with confidence $\psi_t \in \mathbb{R}^{137}$ using OpenPose~\cite{cao2017realtime} including body, hands and facial landmarks.
Additionally, we extract SMPL-X body shape $\boldsymbol{\beta} \in \mathbb{R}^{10}$ and body pose $\boldsymbol{\theta}_t \in \mathbb{R}^{52 \times 3}$, as well as camera parameters $\boldsymbol{\pi}_t = [\mathbf{K}_t \in \mathbb{R}^{3 \times 3}, \mathbf{E}_t \in \mathbb{SE}(3)]$ using SMPLer-X~\cite{cai2024smpler}.

We find that high quality alignment between the reprojected SMPL-X model and human pixels is crucial to final results.
Indeed, most previous works~\cite{chen2021animatable,jiang2023instantavatar,lei2023gart,hu2024gaussianavatar} jointly refine SMPL/SMPL-X parameters along reconstructing the human avatar.
However, we find it sufficient to refine SMPL-X in the preprocessing step, akin to SMPLify-X~\cite{pavlakos2019expressive}, without joint optimizing the avatar.
Concretely, we seek to solve the following optimization problem that balances between pixel alignment and temporal smoothness:
\begin{equation}
\begin{aligned}
    \min_{\boldsymbol{\beta}, \{\boldsymbol{\theta_t}\}, \{\mathbf{b}_t\}}
    \lambda_\text{data} E_\text{data} &+ \lambda_\text{smooth} E_\text{smooth} + \lambda_\text{preserve} E_\text{preserve},
    \\
    E_\text{data} &= \psi_t \rho(\mathbf{k}_t - \hat{\mathbf{k}}_t),
    \\
    E_\text{smooth} &= \| \boldsymbol{\Theta}_{t-1}^T \boldsymbol{\Theta}_t \|,\quad \boldsymbol{\Theta}_t = \texttt{Rodrigues}(\boldsymbol{\theta}_t),
    \\
    E_\text{preserve} &= \|\boldsymbol{\beta} - \boldsymbol{\beta}^{(0)}\| + \|\boldsymbol{\theta}_t - \boldsymbol{\theta}_t^{(0)}\|,
\end{aligned}
\vspace{1em}
\end{equation}
where $\rho$ is the robust Geman-McClure function~\cite{geman1987statistical}, $\hat{\mathbf{k}}_t$ is the reprojected SMPL-X keypoints from current estimates, and $\boldsymbol{\beta}^{(0)}, \boldsymbol{\theta}_t^{(0)}$ are the initial SMPL-X prediction.
We set $\lambda_\text{data}= 100.0, \lambda_\text{smooth} = 10000.0, \lambda_\text{preserve} = 60.0$ throughout our experiments.
We optimize with the second order LBFGS optimizer~\cite{liu1989limited} with a learning rate $\eta = 1.0$ for a total epochs $K = 40$.

\subsection{Globally-consistent surfel avatar}
\label{sec:avatar}
We encode the human appearance and geometry with a global set of 3D Gaussian surfels~\cite{dai2024high} that allows expressive differentiable rendering and surface modeling.
Similar to existing Gaussian-based avatars~\cite{lei2023gart, hu2024gaussianavatar, zielonka2023drivable}, we define surfels in a single canonical space, which can be reposed using forward skinning as opposed to backward root-finding used in previous NeRF-based avatars~\cite{chen2021snarf, li2022tava, jiang2023instantavatar}.
A pictorial illustration is shown in Figure~\ref{fig:system}.

\paragraph{Canonical representation.}
For each surfel $\mathbf{g}_0$ that lives in the canonical frame $t_0$,  we define their attributes as
\begin{equation}
    \mathbf{g}_0 \equiv (\boldsymbol{\mu}_0, \mathbf{R}_0, s, \mathbf{c}, \tau),
\end{equation}
where the position $\boldsymbol{\mu}_0 \in \mathbb{R}^3$ and the orientation $\mathbf{R}_0 \in \mathbb{SO}(3)$ can be reposed, the scale $s \in \mathbb{R}$ and the color $\mathbf{c} \in \mathbb{R}^3$ are constant across poses.
Additionally, we assign the occlusion $\tau \in [0, 1]$ to each canonical surfel, with $1$ indicating full occlusion, for evaluation.
Similar to GaussianAvatar~\cite{hu2024gaussianavatar}, we treat each surfel as an oriented round disk with isotropic scale to prevent needle-like artifacts after reposing.
We keep surfels constantly opaque,~\textit{i.e.} $o = 1$, to avoid semi-transparent surfaces after alpha compositing.
The surfel normal $\mathbf{n}_0$ can be read out trivially as the last column component in $\mathbf{R}_0$.

We find that explicit parameters tend to have large variance after convergence given sparse supervision, which leads to high frequency artifacts when applying score distillation sampling~\cite{poole2022dreamfusion}.
To this end, we employ a hybrid parameterization of surfel attributes.
Specifically, we define $\boldsymbol{\mu}_0$ and $\mathbf{R}_0$ as explicit parameters and use a hash-based MLP network $\Phi$ to predict $s$ and $\mathbf{c}$:
\begin{equation}
    \Phi: \boldsymbol{\mu}_0 \mapsto s, \mathbf{c},
\end{equation}
where each attribute has its own shallow MLP network, taken as input a shared hash grid encoding~\cite{muller2022instant}.
We ablate over this design choice in Section~\ref{sec:ablation}.

We initialize surfels in a predefined virtruvian pose by subdividing corresponding SMPL-X mesh.
Concretely, we subdivide SMPL-X mesh twice and obtain $N=167333$ oriented vertices, which is used to initialize $\boldsymbol{\mu}_0, \mathbf{R}_0$.
We compute the initial $s$ as the average point-to-point distance between each surfel and its $3$-nearest neighbors, as per~\cite{kerbl3Dgaussians}.
Since we adopt implicit parameterization $\Phi$ for $s$, we supervise $\Phi$ with our pre-computed $(\boldsymbol{\mu}_0, s)$ labels for proper initialization.

\paragraph{Articulation representation.}
Given the SMPL-X parameters $\boldsymbol{\beta}, \boldsymbol{\theta}_t$, we can compute their corresponding bone transformations $\{\mathbf{B}_{t,j}\}$ for each joints $j$.
We then articulate each canonical surfel $\mathbf{g}_0$ to posed surfel $\mathbf{g}_t \equiv (\boldsymbol{\mu}_t, \mathbf{R}_t, s, \mathbf{c})$ by linear blend skinning
\begin{equation}
    [\mathbf{R}_t | \boldsymbol{\mu}_t] = \mathbf{B}_t \cdot [\mathbf{R}_0 | \boldsymbol{\mu}_0],\quad
    \text{where }
    \mathbf{B}_t = \sum_j w_j \mathbf{B}_{t,j}.
\end{equation}
$w_j$ is the average skinning weight of the nearest $K = 30$ SMPL-X vertices, weighted by the point-to-point distances in canonical space, similar to~\cite{guo2023vid2avatar,hu2024gaussianavatar}.

We note that, this articulation formulation is much simpler than previous NeRF-based approach that uses backward root-finding~\cite{chen2021snarf,li2022tava,jiang2023instantavatar}.
By adopting forward skinning, our method naturally supports out-of-distribution reposing.

\paragraph{Rendering.}
Each posed surfel $\mathbf{g}_t$ can be efficiently rasterized onto the image plane based on camera parameters $\boldsymbol{\pi}_t$.
For example, RGB image $\hat{\mathbf{I}}_t$ can be rendered by
\begin{equation}
    \hat{\mathbf{I}_t}(\mathbf{x}) = \sum_{i \in \mathcal{H}_t(\mathbf{x})} T_i \alpha_i \cdot \mathbf{c}_i,
\end{equation}
where $T_i$ and $\alpha_i$ are the transmittance and opacity of each projected 2D Gaussian surfel. $\mathcal{H}_t(\mathbf{x})$ is the set of surfels that intersect the ray originated from pixel $\mathbf{x}$.
We can render mask $\hat{\mathbf{M}}_t$, depth map $\hat{\mathbf{D}}_t$, normal map $\hat{\mathbf{N}}_t$, and occlusion map $\hat{\mathbf{O}}_t$ similarly.
This process is fully differentiable and allows end-to-end training from 2D observations.

\subsection{Initial reconstruction}
\label{sec:reconstruction}
We start our optimization process by initial reconstruction, while reasoning about 3D occlusion of our model with respect to the input views.

We adopt both image supervision and structural priors from our preprocessed data for optimization.
During each training iteration, we randomly sample a training view with its corresponding camera and SMPL-X parameters.
Concretely, we seek to solve the following optimization problem:
\begin{equation}
% \begin{gathered}
    \begin{aligned}
\min_{\{\boldsymbol{\mu}_0\}, \{\mathbf{R}_0\}, \Phi} & L_\text{rgb} + \lambda_\text{mask} L_\text{mask} + \lambda_\text{normal} L_\text{normal} + L_\text{reg}, \\
        L_{\text{rgb}} &= 0.2 \cdot \|\mathbf{I}_t - \hat{\mathbf{I}}_t\|_1 + 0.8 \cdot \texttt{SSIM}(\mathbf{I}_t, \hat{\mathbf{I}}_t) + \texttt{LPIPS}(\mathbf{I}_t,  \hat{\mathbf{I}}_t), \\
        L_{\text{mask}} &= \|\mathbf{M}_t - \hat{\mathbf{M}}_t\|_1, \\
        L_{\text{normal}} &= l_\text{normal}(\mathbf{N}_t, \hat{\mathbf{N}}_t) + l_\text{normal}({}^B\mathbf{N}_t, {}^B\hat{\mathbf{N}}_t) \\
        l_{\text{normal}}(\mathbf{N}, \hat{\mathbf{N}}) &= 0.2 \cdot \mathbf{N}^T \hat{\mathbf{N}} + \texttt{LPIPS}(\mathbf{N}, \hat{\mathbf{N}}).
        % \end{gathered}
    \end{aligned}
% \end{gathered}
\label{eq:reconstruction}
\end{equation}
Similar to TeCH~\cite{huang2023tech}, we find that LPIPS~\cite{zhang2018unreasonable} works with normal supervision and encourages crisp geometry over overly smoothed solution.
To render back normal $^B\hat{\mathbf{N}}_t$, we rasterize by sorting surfels in descending depth order as opposed to usual ascending order.
Using back normal supervision, the geometry of our avatar is constrained in unobserved regions.

We set $\lambda_\text{mask} = 1.0$ and $\lambda_\text{normal} = 1.0$ throughout our experiments. 

Our regularization term $L_\text{reg}$ consists of normal-depth consistency loss and curvature loss from~\cite{dai2024high}, as well as an offset and scale regularization from~\cite{hu2024gaussianavatar} that penalizes irregular solution.
This reconstruction process is trained with an Adam optimizer~\cite{kingma2014adam} for a total steps $K = 500$.
The whole process takes about 5 minutes to finish.

As a side task, we are interested in estimating occlusion of our human model during the optimization process for quantify the portion of human body that has been observed from input video.

This is achieved by optimizing occlusion map $\hat{\mathbf{O}}_t$ in each training view per iteration, \textit{i.e.},
\begin{equation}
    \min_{\{\tau\}} \|\hat{\mathbf{O}}_t\|_1.
\end{equation}
Note that we detach all gradient from this objective towards other surfel properties such that we are only estimating the self-occlusion of our \textit{current} geometry with respect to training views, without affecting the reconstruction process.
We find that it is necessary to perform back-face culling~\cite{dai2024high} when rendering occlusion map.
Without this operation, the occlusion signal "leaks" onto the back of the human figure. 

\subsection{Generative refinement}
\label{sec:generation}
After initial reconstruction and occlusion estimation, we have a partially completed avatar.
We next refine the initial result by score distillation sampling~(SDS) a diffusion model~\cite{poole2022dreamfusion}.

In this work, we use ImageDream~\cite{wang2023imagedream} as our diffusion prior.
Empirically, we find this image-conditional multi-view diffusion model to be much more reliable compared to other alternatives, such as MVDream~\cite{shi2023mvdream} or SD~\cite{rombach2022high}.
These alternatives rely heavily on text prompt and often produce overly saturated textures that are inconsistent with the original video.
For example, TeCH~\cite{huang2023tech} needs to finetune a SD model.

In addition to the set of losses in Equation~\ref{eq:reconstruction}, we sample\footnote{In practice we sample $4$ views for~\cite{wang2023imagedream} and discuss one-view rendering for simplicity.} a novel view camera $\tilde{\boldsymbol{\pi}}$ and render novel view $\tilde{\mathbf{I}}_t, \tilde{\mathbf{N}}_t$ during each training iteration for SDS supervision using the SMPL-X parameter in the current batch. 
The diffusion process is conditioned on image prompts $\mathbf{I}_t$, $\mathbf{N}_t$ and text prompt $\mathbf{p}$,
\begin{equation}
\begin{gathered}
\min_{\{\boldsymbol{\mu}_0\}, \{\mathbf{R}_0\}, \Phi}
\lambda_\text{rgb}^\text{sds} L_\text{rgb}^\text{sds} + \lambda_\text{normal}^\text{sds} L_\text{normal}^\text{sds},
\\
\begin{aligned}
    L_\text{rgb}^\text{sds} &=
    \mathbb{E}_{i, \epsilon}
    \Big[ 
    \big\| \tilde{\mathbf{I}}_t - \texttt{Denoise}_\Psi(\tilde{\mathbf{I}}_t; \mathbf{I}_t, \mathbf{p}, i, \epsilon) \big\|_2^2 
    \Big],
    \\
    L_\text{normal}^\text{sds} &=
    \mathbb{E}_{i, \epsilon}
    \Big[ 
    \big\| \tilde{\mathbf{N}}_t - \texttt{Denoise}_\Psi(\tilde{\mathbf{N}}_t; \mathbf{N}_t, \mathbf{p}, i, \epsilon) \big\|_2^2 
    \Big],
\end{aligned}
\end{gathered}
\label{eq:generation}
\end{equation}
where $\texttt{Denoise}_\Psi$ denotes the a full denoising step from timestep $i$ to $0$ using noise $\epsilon$ with pretrained parameter $\Psi$.
Please refer to ImageDream~\cite{wang2023imagedream} for more detail. 
We first refine the shape for $K = 500$ iterations by setting $\lambda_{\text{rgb}}^{\text{sds}} = 0, \lambda_{\text{normal}}^{\text{sds}} = 10^{-4}$.
We then refine the texture for $K=1000$ iterations by setting $\lambda_{\text{rgb}}^{\text{sds}} = 10^{-4}, \lambda_{\text{normal}}^{\text{sds}} = 0$.
The whole process takes about 20 minutes to finish.

\section{Experiments}
Our method is unique in being able to reconstruct self-occluded human from a single video.
We first compare with HAVE-FUN~\cite{yang2024have} on its own benchmark.
After carefully examining the actual occlusion in its evaluation, we find that large portion ($^{\sim}90\%$) of human body is observed during training.
We then devise our own experimental setup to rigorously evaluate the performance of our approach on self-occluded videos, both quantitatively and qualitatively, discussed next.

\subsection{Experimental setup}
\begin{figure}
    \centering
    \includegraphics[width=\linewidth]{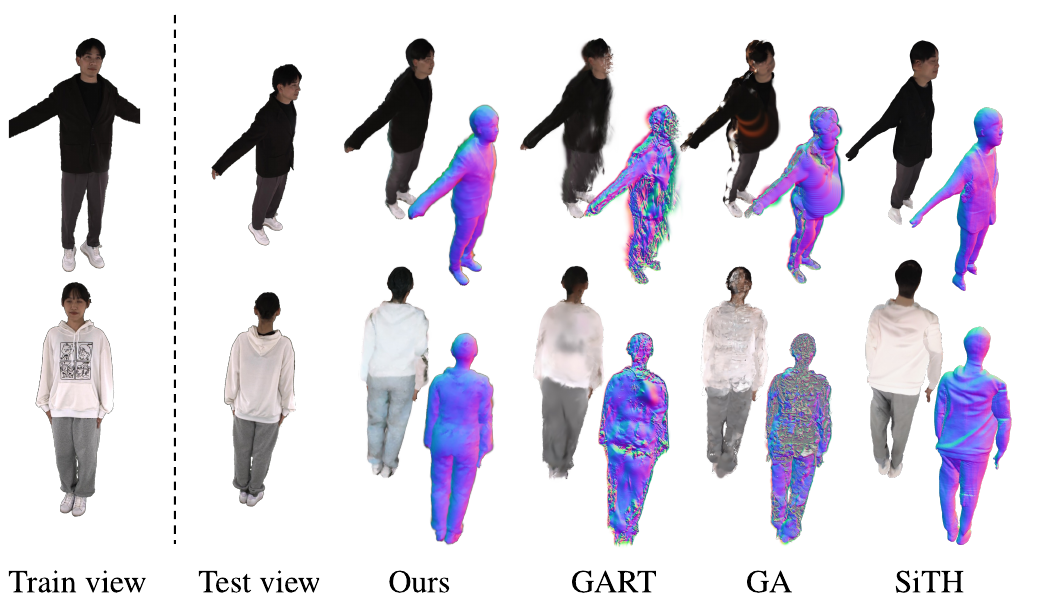}
    \caption{
    \textbf{Qualitative results on DNA-Rendering dataset.}
    For each training view, we visualize the ground-truth novel view along with predicted RGB rendering and normal map from different approaches.
    Our method recovers photo-realistic and geometrically plausible avatars comparing to baselines.
    For GART and GA, we read out their normals by depth gradient~\cite{dai2024high, huang20242d, jiang2023gaussianshader}.
    }
    \label{fig:dna-rendering}
    % \vspace{-1em}
\end{figure}

% \providecommand\animage{}
% \renewcommand{\animage}[2]{
%     \frame{\includegraphics[width=\linewidth,clip,trim=#1]{figures/#2}}
% }
% \begin{figure}[t!]
%     \setlength{\tabcolsep}{0.8pt}
%     \renewcommand{\arraystretch}{0.5}
%     \begin{tabularx}{\textwidth}{@{}*{6}{C}@{}}
%          \animage{0 0 0 0}{train_gt.png} &
%         \animage{0 0 0 0}{gt.png} &
%         \animage{0 0 0 0}{ours.png} &
%         \animage{0 0 0 0}{gart.png} &
%         \animage{0 0 0 0}{ga.png} &
%         \animage{0 0 0 0}{ga.png}
%         \\
%         {Train View} &
%         {GT Reference} &
%         {Ours} &
%         {GART~\cite{TODO}} &
%         {GA~\cite{TODO}} &
%         {SiTH~\cite{TODO}}
%     \end{tabularx}
%     \vspace{-2pt}
%     \caption{
%         \textbf{Qualitative results on DNA-Rendering dataset.} The leftmost image in each row shows the training view at the same time step as the validation view. The regions highlighted in green indicate areas excluded from evaluation due to the lack of co-visibility between training and validation views.
%     }
%     \vspace{-1em}
%     \label{fig: novel view comparison}
% \end{figure}

\paragraph{Datasets.}
While we primarily focus on reconstructing self-occluded humans in-the-wild, it is unpractical to quantitatively evaluate solely based on internet footage. 
Therefore, we show results on three types of dataset: FS-XHuman used by HAVE-FUN~\cite{yang2024have}, a re-purposed multi-view human dataset and a set of internet footage of moving people.
Concretely, we follow HAVE-FUN's evaluation split, which consists of 20 different subjects.
We evaluate over few-view reconstruction given 2 views, 4 views and 8 views, respectively.
After closer look, we find that FS-XHuman has very few occlusion and thus propose our own evaluation on DNA-Rendering dataset~\cite{cheng2023dna} due to its diversity and capture quality.
DNA-Rendering comes with ground truth camera and SMPL-X annotations, making it suitable for fair comparisons between different approaches.
Finally, we experiment with a set of in-the-wild videos from internet. 
These videos feature severe self-occlusion, fast motion and motion blur, making them much harder to reconstruct compared to the ones captured from a light stage.
We use them to demonstrate the robustness of our method in the real-world scenario.
\paragraph{Metrics.}
For quantitative assessment, we evaluate novel view rendering on the FS-XHuman, DNA-Rendering and show qualitative comparisons on in-the-wild videos.
Concretely, we evaluate standard PSNR, SSIM and LPIPS metrics from neural rendering literatures.
In addition to that, we evaluate the rendering quality in occluded regions using mPSNR and mLPIPS~\cite{gatys2017controlling, huh2020transforming}.
This evaluation shed light on how different approaches balance between reconstruction and generation.
Since no ground-truth occlusion map exists in DNA-Rendering dataset, we use the inferred occlusion from our model for this task.
We also propose a new metric called Body Occlusion Ratio~(BOR) to quantify the portion of human body being seeing during training.
It is computed by averaging inferred per-surfel occlusion~$\tau$ for each training sequence.
\paragraph{Baselines.}
We consider both reconstruction-based and generation-based methods as baseline.
For reconstruction-based method, we evaluate against state-of-the-art Gaussian-based avatars including GART~\cite{lei2023gart} and GaussianAvatar~\cite{hu2024gaussianavatar}~(GA).
For generation-based method, there exists no method that can handle video input.
We therefore compare against recent human-specific image-to-3D approach SiTH~\cite{ho2023sith} out of all candidates~\cite{huang2023tech,zhang2023sifu,albahar2023single} due to its efficiency and code availability.
For these methods, we run baselines on three randomly selected frame independently and repose using the input SMPL-X parameters.
As one can expect, they are temporally inconsistent and have severe reposing artifacts due to the inability to properly handle articulation, which we show in Figure~\ref{fig:dna-rendering-consistent}.
The final quantitative metrics are averaged across these independently generated avatars.
\subsection{Results on FS-XHumans dataset}
To compare with concurrent work HAVE-FUN~\cite{yang2024have} that combines reconstruction and generation, we evaluate our method on FS-XHumans.
Quantitative results are shown in Table~\ref{tab:fs-xhuman} and qualitative results are included in supplement.
Our method consistently outperforms HAVE-FUN in terms of PSNR and SSIM while on-par with it for LPIPS metric.
We evaluate BOR for occlusion assessment in Table~\ref{tab:bor}.
As demonstrated, around $^{\sim}90\%$ human body is observed in FS-XHumans, even though its evaluation tries to work in few-view setting.
Comparing to it, we propose a testing split from DNA-Rendering, which aligns closer to in-the-wild videos that have self-occlusion.
\subsection{Results on DNA-Rendering dataset}
\label{sec:dna-rendering}
DNA-Rendering dataset contains $500$ human captures in a light stage setup.
We choose $7$ sequences without object interaction and loose clothing for our experiments.
For each video, we train from a single camera and evaluate novel view rendering from $4$ unseen cameras.
We select the training camera as the one that the human is facing to in the first frame.
With this simple rule, we already make sure that there are parts of human body remaining self-occluded throughout the video because the actors rarely orient on this dataset.
For validation camera selection, we uniformly sample from the provided $60$ cameras such that unobserved regions have ground-truth pixels.
% For more details on this dataset, please refer to our supplementary material.

% \begin{wrapfigure}{r}{0.4\linewidth}
%     \vspace{-2em}
\begin{figure}
    \centering
    \includegraphics[width=\linewidth]{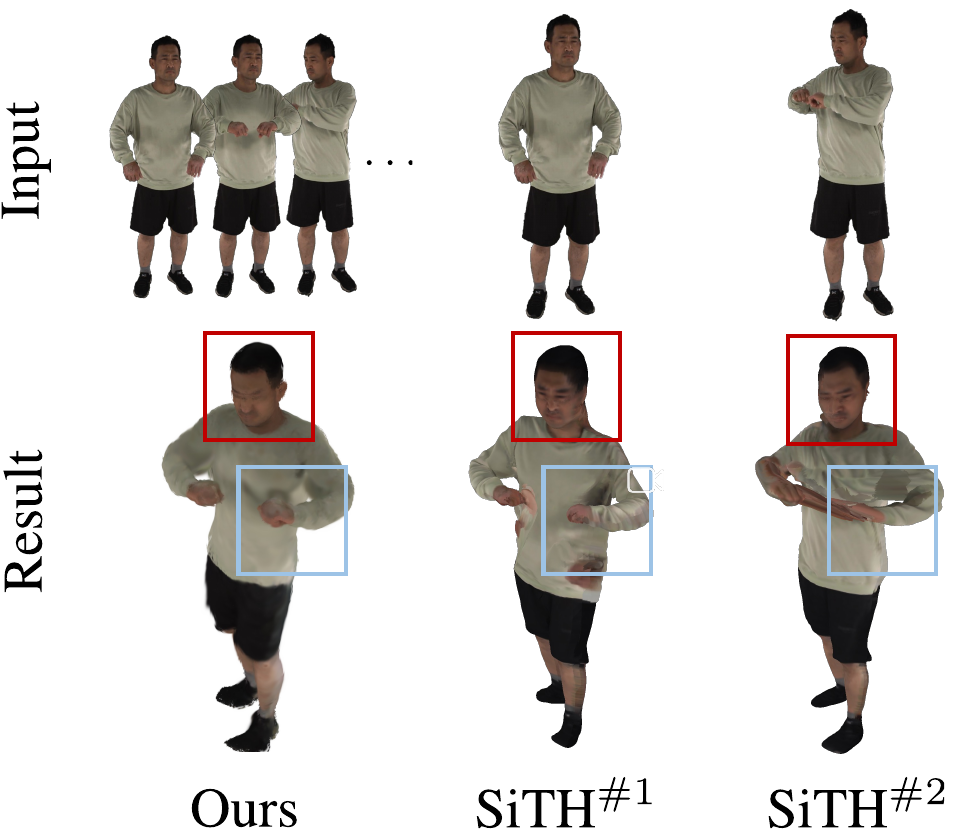}
    \caption{\textbf{Comparison between our globally consistent avatar and image-to-3D baseline.}
    Our method is able to fuse all observations from a video and allow natural reposing.
    }
    \label{fig:dna-rendering-consistent}
\end{figure}
%     \vspace{-1em}
% \end{wrapfigure}

We report our quantitative results in Table~\ref{tab:dna-rendering}.
Our proposed method outperforms all baselines on all metrics by a substantial margin.
We separate our baselines into two categories: reconstruction-based methods GART and GA, generation-based method SiTH.
When evaluating against the full image, denoted as ``Full'', our method improves significantly over baselines, with +1.2 PSNR and -10\% LPIPS improvements comparing to reconstruction-based methods.
This improvement is even larger comparing to generation-based baseline SiTH in terms of PSNR by +5.3, with similar -10\% LPIPS improvement.
It is perhaps not surprising giving that PSNR favors exactness over realism while LPIPS does the other way around.
It is also notable that our approach dramatically improves over the existing approaches in both visible and occluded regions (noted as ``Visible'' and ``Occlusion'' in the table).

\begin{table*}[t]
\centering
\tablestyle{6.5pt}{1.1}
\resizebox{0.8\linewidth}{!}{ 
\begin{tabular}{lccccccc}
\toprule
\multirow{2}{*}{\textbf{Method}} & \multirow{2}{*}{\textbf{Type}} & \multicolumn{2}{c}{\textbf{Full}} & \multicolumn{2}{c}{\textbf{Visible}} & \multicolumn{2}{c}{\textbf{Occlusion}} \\
\cmidrule(lr){3-4}
\cmidrule(lr){5-6}
\cmidrule(lr){7-8}
& & PSNR$\uparrow$ & LPIPS$\downarrow$ & mPSNR$\uparrow$ & mLPIPS$\downarrow$ & mPSNR$\uparrow$ & mLPIPS$\downarrow$\\
\midrule
SiTH~\cite{ho2023sith} & Gen. & 17.84 & 0.065 & 14.55 & 0.282 & 9.63 & 0.271 \\
\midrule
GA~\cite{hu2024gaussianavatar} & \multirow{2}{*}{Recon.} & 21.97 & 0.068 & 15.28 & 0.456 & 10.06 & 0.492 \\
GART~\cite{lei2023gart} & & 21.89 & 0.067 & 16.33 & 0.290 & 12.86 & 0.283 \\
\midrule
Ours & Gen. + Recon. & \textbf{23.16} & \textbf{0.055} & \textbf{19.96} & \textbf{0.227} & \textbf{14.54} & \textbf{0.253} \\
\bottomrule
\end{tabular}%
}
\vspace{1em}
\captionof{table}{
\textbf{Quantitative results on DNA-Rendering dataset}. 
We evaluate the novel view rendering performance of all approaches in full image (``Full''), visible regions (``Visible'') and occluded regions (``Occlusion'').
Our method consistently out-performs different baselines in all metrics by a significant margin.
Best metrics are marked as bold.
}
\label{tab:dna-rendering}
\end{table*}
\begin{table}[t]
\centering
\tablestyle{6.5pt}{1.1}
\resizebox{\linewidth}{!}{ 
\begin{tabular}{lccccccc}
\toprule
\textbf{Method} & \textbf{\#Input} & PSNR$\uparrow$ & SSIM$\uparrow$ & LPIPS$\downarrow$\\
\midrule
\multirow{2}{*}{SelfRecon~\cite{jiang2022selfrecon}} & 8-shots & 19.90 & 0.927 & 0.065 \\
 & 100-shots & 20.70 & 0.943 & 0.063 \\
\midrule
TeCH~\cite{huang2023tech} & 1-shot & 21.00 & 0.924 & 0.065 \\
\midrule
\multirow{3}{*}{HaveFun~\cite{yang2024have}} & 2-shots & 24.00 & 0.955 & \textbf{0.042} \\
 & 4-shots & 25.60 & 0.963 & \textbf{0.035} \\
 & 8-shots & 26.80 & 0.967 & \textbf{0.030} \\
\midrule
% Ours & Gen. + Recon. & \textbf{23.16} & \textbf{0.055} & \textbf{19.96}  \\
\multirow{3}{*}{Ours} & 2-shots & \textbf{25.18} & \textbf{0.958} & 0.049 \\
 & 4-shots & \textbf{26.86} & \textbf{0.964} & 0.043 \\
 & 8-shots & \textbf{27.94} & \textbf{0.968} & 0.039 \\
\bottomrule
\end{tabular}%
}
\vspace{1em}
\captionof{table}{
\textbf{Comparison on FS-XHumans dataset}. 
% We evaluate the novel view rendering performance of all approaches in full image (``Full''), visible regions (``Visible'') and occluded regions (``Occlusion'').
% Our method consistently out-performs different baselines in all metrics by a significant margin.
% Best metrics are marked as bold.
We compare our methods with few-shot human reconstruction methods. Our method consistently outperforms different baselines in PSNR and SSIM metrics by a significant margin. Best metrics in 2-view/4-view/8-view settings are marked as bold.
}
\label{tab:fs-xhuman}
\end{table}
\begin{table}[t]
\centering
\tablestyle{19.5pt}{1.1}
\resizebox{1.0\linewidth}{!}{ 
\begin{tabular}{lccccccc}
\toprule
\textbf{Dataset} & \textbf{BOR} \\
\midrule
% NeuMan~\cite{jiang2022neuman} & 0.159\\
FS-XHumans~\cite{yang2024have} & 0.202/0.138/0.095 \\
% Occ-ZJU-MoCap~\cite{Xiang_2023_OccNeRF}  & 0.181 \\
DNA-Rendering~\cite{cheng2023dna} & 0.460\\
In-the-wild & 0.341\\
\bottomrule
\end{tabular}%
}
\vspace{1em}
\captionof{table}{
\textbf{BOR value of different datasets}. 
\looseness=-1
FS-XHumans have 2-view/4-view/8-view evaluation.
However, only around $^{\sim}10\%$ of human body remains unobserved.
In comparison, our proposed DNA-Rendering split aligns closer to self-occluded videos in-the-wild.
% We evaluate the novel view rendering performance of all approaches in full image (``Full''), visible regions (``Visible'') and occluded regions (``Occlusion'').
% Our method consistently out-performs different baselines in all metrics by a significant margin.
% Best metrics are marked as bold.
}
\label{tab:bor}
\end{table}
We visualize our qualitative comparisons in Figure~\ref{fig:dna-rendering} with both novel view RGB rendering and normal map predictions.
For GART and GA, we read out their normals by depth gradient~\cite{dai2024high, huang20242d, jiang2023gaussianshader}.
We want to emphasize our improvements in three aspects.
First, our method produces crisp geometry details as suggested by predicted normal maps.
Second, it is clear to see that our method produces highly realistic synthesis on the unobserved regions, \textit{e.g.}, around the back regions in the second row.
Reconstruction-based methods struggle in these under-constrained areas.
Third, our reconstruction component helps prevent unnatural shapes -- this is evident in the first row where SiTH produces very thin arms.
\looseness=-1

Finally, we demonstrate that our approach benefits from a globally consistent representation for avatar creation such that we can fuse observations from multiple video frames.
We visualize our reposed avatar in Figure~\ref{fig:dna-rendering-consistent}, comparing to two different runs by SiTH, which can only condition on a single input frame.
The note three observations.
First, fusing observations from multiple frames by reconstruction resolves ambiguity in seen regions (in blue bounding box), where SiTH strugle to produce accurate shape solely based on image prior.
Second, SiTH produces inconsistent results between runs as shown in the red bounding box.
Third, our globally consistent avatar representation allow much more natural reposing comparing to SiTH mesh skinning. 

\begin{figure*}
    \centering
    \includegraphics[width=\linewidth]{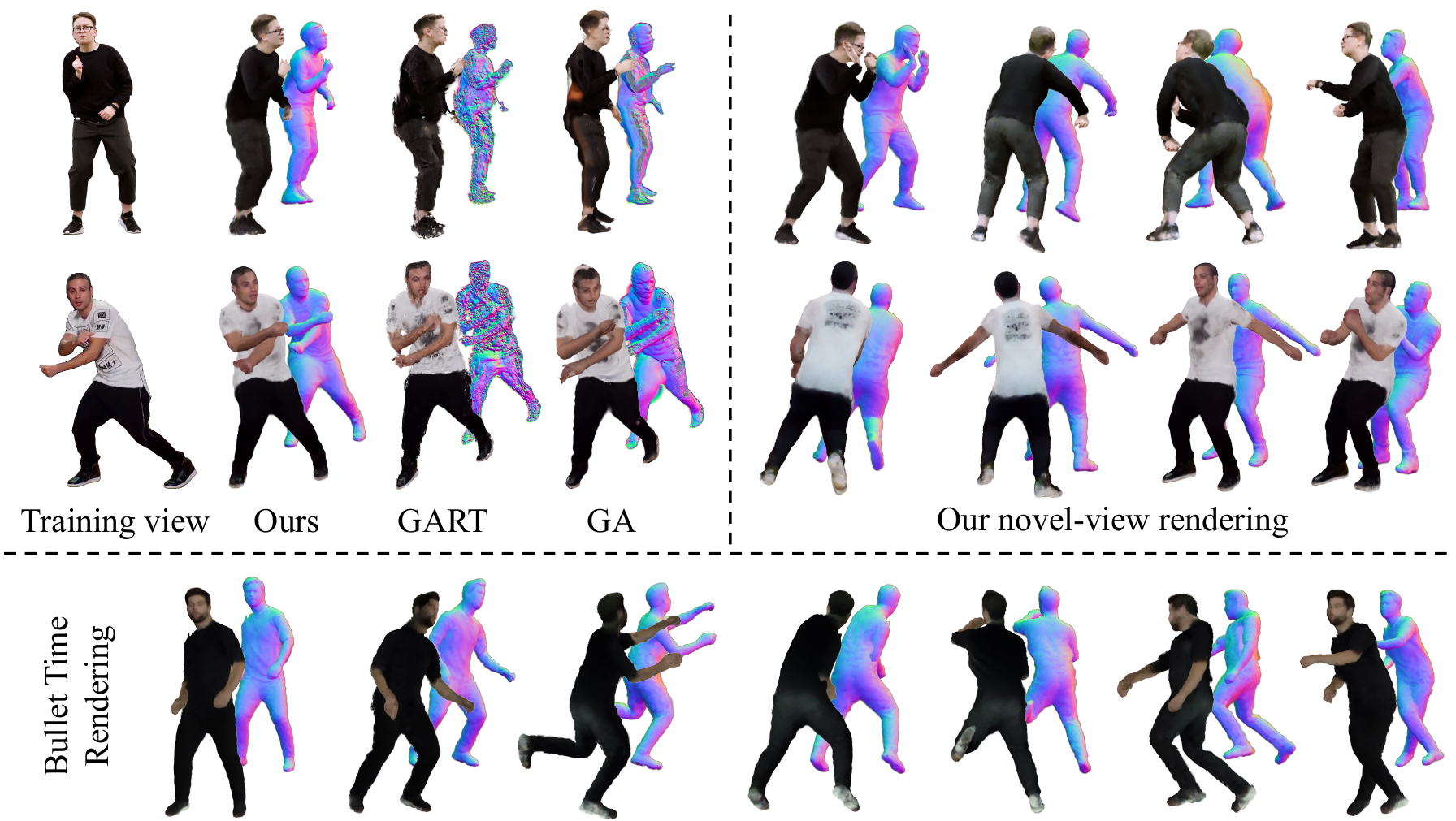}
    \caption{
    \textbf{Qualitative results on in-the-wild videos.}
    We visualize novel-view rendering comparison in top left, our 360 rendering on top right, and our bullet time rendering on the bottom.
    We visualize both the RGB rendering and normal map rendering in each result.
    }
    \label{fig:in-the-wild}
\end{figure*}
\subsection{Results on in-the-wild videos}
We report the qualitative results of our method applied to in-the-wild videos, as shown in Figure \ref{fig:in-the-wild}.
Our dataset comprises single human-centered internet videos with severe self-occlusion. 
The first row of the figure displays the novel view rendering results.
We conducted comparisons with methods GART and GA, which often fail to accurately reconstruct human shape and texture under these challenging conditions.
In contrast, our method consistently produces high-detail normal maps and realistic textures.
Additionally, we include our reposing results to further demonstrate the robustness of our approach.
Our method again is able to produce photo-realistic rendering and accurate shape prediction.
\looseness=-1

\subsection{Ablation}
\label{sec:ablation}
\begin{figure}
    % \vspace{-3em}
    \centering
    \includegraphics[width=\linewidth]{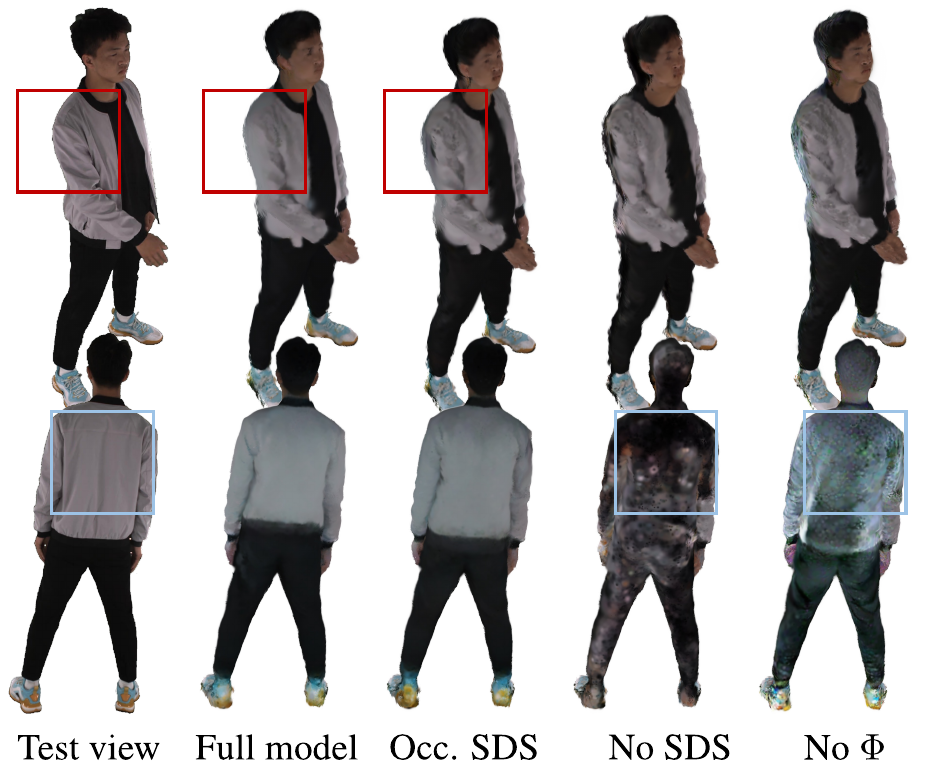}
    \captionof{figure}{\textbf{Ablation.}
    We ablate over occlusion masking in SDS, generation itself and our implicit parameterization $\Phi$.
    }
    \label{fig:ablation}
    % \vspace{-3em}
\end{figure}
\begin{table}[t]
\centering
\tablestyle{6.5pt}{1.1}
\resizebox{\linewidth}{!}{ 
\begin{tabular}{lccccccc}
\toprule
\multirow{2}{*}{\textbf{Method}} & \multicolumn{2}{c}{\textbf{Full}} & \multicolumn{2}{c}{\textbf{Visible}} & \multicolumn{2}{c}{\textbf{Occlusion}} \\
\cmidrule(lr){2-3}
\cmidrule(lr){4-5}
\cmidrule(lr){6-7}
& PSNR$\uparrow$ & LPIPS$\downarrow$ & mPSNR$\uparrow$ & mLPIPS$\downarrow$ & mPSNR$\uparrow$ & mLPIPS$\downarrow$\\
\midrule
W/o sds & 21.47 & 0.065 & 17.14 & 0.275 & 9.78 & 0.441 \\
W/o implicit & 22.86 & 0.066 & 17.74 & 0.275 & 11.01 & 0.455 \\
\midrule
Ours w/ occ & \textbf{23.58} & \textbf{0.055} & 17.91 & 0.256 & \textbf{12.15} & \textbf{0.324} \\
Ours & 23.51 & \textbf{0.055} & \textbf{18.28} & \textbf{0.247} & 11.93 & \textbf{0.324} \\
\bottomrule
\end{tabular}%
}
\vspace{1em}
\captionof{table}{
\textbf{Ablation results on DNA-Rendering dataset}. 
We ablate the novel view rendering performance of all approaches in full image (``Full''), visible regions (``Visible'') and occluded regions (``Occlusion'').
Our method consistently out-performs different baselines in all metrics by a significant margin.
Best metrics are marked as bold.
}
\label{tab:ablation}
\end{table}

% \begin{table}[t]
% \centering
% \tablestyle{6.5pt}{1.1}
% \resizebox{\linewidth}{!}{ 
% \begin{tabular}{lccccc}
% \toprule
% \multirow{2}{*}{\textbf{Method}} & \multicolumn{2}{c}{\textbf{Full}} & \multicolumn{1}{c}{\textbf{Visible}} & \multicolumn{1}{c}{\textbf{Occlusion}} \\
% \cmidrule(lr){2-3}
% \cmidrule(lr){4-4}
% \cmidrule(lr){5-5}
% & PSNR$\uparrow$ & LPIPS$\downarrow$ & mLPIPS$\downarrow$ & mLPIPS$\downarrow$\\
% \midrule
% W/o sds & 21.47 & 0.065 & 0.275 & 0.441 \\
% W/o implicit & 22.86 & 0.066 & 0.275 & 0.455 \\
% \midrule
% Ours w/ occ & \textbf{23.58} & \textbf{0.055} & 0.256 & \textbf{0.324} \\
% Ours & 23.51 & \textbf{0.055} & \textbf{0.247} & \textbf{0.324} \\
% \bottomrule
% \end{tabular}%
% }
% \vspace{1em}
% \captionof{table}{
% \textbf{Ablation results}. 
% Best metrics are marked as bold.
% }
% \label{tab:ablation}
% \end{table}
We conducted ablation study about our design choices and consider following baselines:
(1) our model with occlusion masking in SDS loss as discussed in Section~\ref{sec:generation} (``Occ. SDS''); 
(2) our model without the generation component (``No SDS'');
(3) our model without implicit parameterization $\Phi$ (``No $\Phi$'').
Qualitative results are shown in Figure~\ref{fig:ablation} and quantitative results are included in Table~\ref{tab:ablation}. Each component is beneficial and contributes to our full model.
\looseness=-1

\section{Discussion and Conclusion}
While we present promising steps towards robust human avatar recovery from in-the-wild videos several limitations remain.  
It inherits the issue of generating saturated colors from SDS-based methods, remains a test-time optimization approach limiting interactive use, and lacks a comprehensive in-the-wild dataset with ground-truth multi-view annotations for better evaluation. 
Future work includes training human-specific multi-view diffusion models on large-scale human capture data and creating an in-the-wild human dataset with multi-view validation.
Despite these limitations, we presented SOAR for self-occluded avatar recovery from a single in-the-wild video, employing a globally-consistent surfel model for fusing noisy supervision and reposing, and leveraging structural human normal priors and generative diffusion priors. 
Our method recovers photo-realistic avatar models with plausible shapes, significantly improving over existing methods. 
Experiments on multi-view datasets and in-the-wild videos demonstrate that our method achieves state-of-the-art performance compared to purely reconstruction-based and generation-based methods.

\section{Acknowledgement}
This project is supported in part by DARPA No. HR001123C0021, IARPA DOI/IBC No. 140D0423C0035, NSF:CNS-2235013, Bakar Fellows, and Bair Sponsors. The views and conclusions contained herein are those of the authors and do not represent the official policies or endorsements of these institutions.

{
    \small
    \bibliographystyle{ieeenat_fullname}
    \bibliography{main}
}

% \input{sec/0_abstract}    
% \input{sec/1_intro}
% \input{sec/2_formatting}
% \input{sec/3_finalcopy}
% {
%     \small
%     \bibliographystyle{ieeenat_fullname}
%     \bibliography{main}
% }
% \input{sec/X_suppl}

\end{document}